\documentclass[10pt]{article}

\usepackage[T1]{fontenc}
\usepackage[utf8]{inputenc}
\usepackage{lmodern}
\usepackage{microtype}
\usepackage{geometry}
\usepackage{graphicx}
\usepackage{booktabs}
\usepackage{adjustbox}
\usepackage{amsmath, amssymb}
\usepackage{siunitx}
\usepackage{xcolor}
\usepackage[hidelinks]{hyperref}
\usepackage[nameinlink,capitalise,noabbrev]{cleveref}

\usepackage{indentfirst}

\usepackage[numbers,sort&compress]{natbib}

\title{\textbf{Cholec80-port: A Geometrically Consistent Trocar Port Segmentation Dataset\\for Robust Surgical Scene Understanding}}

\author{
Shunsuke Kikuchi$^{1}$ \quad Atsushi Kouno$^{1}$ \quad Hiroki Matsuzaki$^{1}$\\[2pt]
$^{1}$Jmees Inc., Kashiwa-city, Chiba, Japan\\
\texttt{engineer@jmees-inc.com}
}

\date{} 

\begin{document}
\maketitle

\vspace{-2mm}

\section{Introduction}
Precise segmentation of surgical instruments and anatomical structures is essential for advanced surgical scene understanding. Beyond recognition, many downstream geometry-based tasks require robust separation of local motion (dynamic objects) from global motion (anatomical background), including image stitching, 3D reconstruction, and visual SLAM (vSLAM). Prior work has shown that masking dynamic regions during feature extraction can improve geometric pipelines by preventing trackers from following moving outliers~\cite{3dtracking}. 

In laparoscopic surgery, the trocar port serves as the physical gateway through the abdominal wall. Because the endoscopic camera passes through these ports, they can obstruct the field of view. While ports are most visible during camera insertion and retraction, they may remain partially visible throughout procedures as peripheral or even central occlusions. Importantly, port surfaces are often specular and textured, attracting an excessive number of feature points. Unlike surgical instruments, ports are approximately camera-fixed and persist over long durations, making them uniquely detrimental to geometry-based methods: they repeatedly introduce strong, non-anatomical features that can bias matching, increase geometric error, and cause alignment artifacts.

Despite this utility, explicit port labels are conspicuously absent from most large-scale datasets. This scarcity is largely attributed to de-identification concerns, because ports can capture transitions to the exterior environment; consequently, ports are frequently merged into broader categories (e.g., ``abdominal wall'' in related datasets).

Currently, only two public datasets, m2caiSeg~\cite{m2caiSeg} and GynSurg~\cite{gynsurg2025}, provide explicit labels for ``trocars'' or ``cannulas''. However, both contain limitations for geometry-aware applications. m2caiSeg is small ($N{=}370$) and exhibits annotation artifacts (e.g., interpolation noise). GynSurg is larger ($N{=}4873$), but its COCO polygon representation can induce a ``hole-filling'' policy in which the central lumen is masked, which is geometrically inconsistent when organs are visible through the opening.
To address these limitations, we propose Cholec80-port\footnote{Datase and Source code is available at \texttt{https://github.com/JmeesInc/cholec80-port} }
Our contributions are:
\begin{enumerate}
    \item A rigorous and practical annotation SOP that prioritizes geometric fidelity by defining a port-sleeve mask that excludes the central opening.
    \item A large-scale dataset from Cholec80 and data cleansing of existing datasets to unify them under the proposed SOP.
    \item A pretrained baseline port segmentation model released with tooling to support downstream surgical computer vision research.
\end{enumerate}

\begin{figure}[h!]
    \centering
    \includegraphics[scale=0.5]{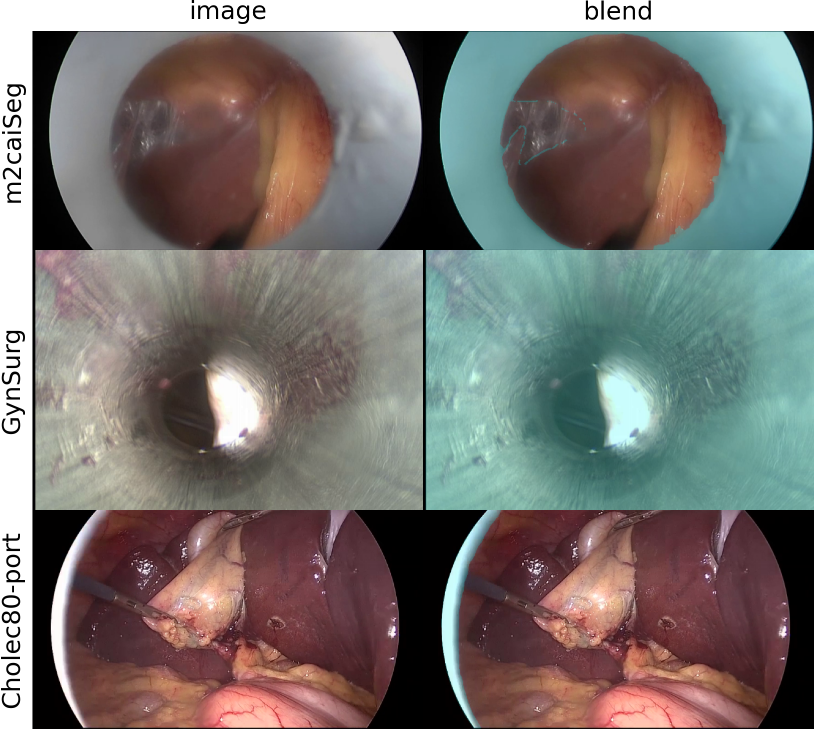}
    \caption{\textbf{Annotation consistency.} Examples of erroneous labels in m2caiSeg (top) and GynSurg (middle) and the proposed geometrically consistent port-sleeve labels in Cholec80-port (bottom). For each row, the original frame (left) and an overlay visualization (right) are shown.}
    \label{fig:annotation_examples}
\end{figure}

\section{Methods}
\subsection{Dataset acquisition and sampling}
We utilized the first 20 videos from Cholec80. To balance surgical scene diversity with annotation efficiency, we sampled every 30 frames, resulting in 38{,}434 annotated frames. The dataset was split at the video level to avoid leakage: Videos 01--08 for training, 09--10 for validation, and 11--20 for testing.

Among sampled frames, 1{,}398 frames contain visible ports. This yields a substantially larger positive sample size than m2caiSeg ($n{=}255$) and GynSurg ($n{=}130$), improving robustness for both segmentation and detection-like use cases.

\subsection{Annotation SOP: port-sleeve definition}
All annotations were performed using CVAT~\cite{cvat}. We define the target region as the sleeve: the rigid metallic or plastic cylindrical component visible beyond the internal valve. Crucially, the sleeve definition \textbf{excludes} the central lumen (opening), because masking the lumen can suppress anatomically valid pixels and introduce geometric inconsistencies for image-plane aggregation and feature extraction.

For ambiguous frames where boundaries are unclear due to orientation or specular highlights, we consulted temporal context (neighboring frames) to confirm the physical extent of the sleeve. This temporal verification helps distinguish sleeve boundaries from transient reflections and motion blur.

\subsection{Cleansing and unification of existing datasets}
We additionally cleanse and unify existing datasets under the same SOP:
\begin{itemize}
    \item m2caiSeg: We re-annotated the dataset to remove interpolation artifacts and spurious masks. After cleansing, only a small subset satisfies the proposed high-quality, sleeve-consistent criteria.
    \item GynSurg: We corrected the hole-filling issue by segmenting central lumen regions and subtracting them from the original polygons to derive sleeve-only masks.
\end{itemize}

\subsection{Model and training protocol}
We evaluate dataset efficacy using a ConvNeXt-Base~\cite{convnext} encoder with a U-Net~\cite{unet} decoder for binary semantic segmentation. We train with a standard combined objective:
\begin{equation}
\mathcal{L} \;=\; \mathcal{L}_{\text{Dice}}(Y_{\text{pred}}, Y_{\text{true}}) \;+\; \mathcal{L}_{\text{BCE}}(Y_{\text{pred}}, Y_{\text{true}}).
\end{equation}
Training uses AdamW~\cite{adamw} with learning rate $5\times10^{-5}$, batch size 16, and input resolution $384\times384$.

\subsection{Evaluation metrics}
We report two complementary metrics:
\begin{enumerate}
    \item Dice score (segmentation fidelity): computed only on frames where ports are present (i.e., $\lVert Y_{\text{true}}\rVert_1 > 0$) to measure boundary recovery when the target exists.
    \item Detect F1 (frame-level robustness): a frame is classified as positive if at least one pixel is predicted/annotated as port, and negative otherwise; we then compute F1 across frames.
\end{enumerate}
To quantify the impact of SOP consistency and cleansing, we perform ablations using original (uncleaned) datasets under the same model and protocol.


\section{Results}
\begin{table}[t]
    \centering
    \caption{\textbf{Port segmentation results (Dice and Detect F1).} Rows denote the training dataset; columns denote the evaluation dataset. Dice is computed only on port-present frames (GT$>0$).}
    \label{tab:results}
    \begin{adjustbox}{max width=\linewidth}
    \begin{tabular}{l cc cc cc}
        \toprule
        & \multicolumn{2}{c}{m2caiSeg (test)} & \multicolumn{2}{c}{GynSurg (Video 09--10)} & \multicolumn{2}{c}{Cholec80-port (test)} \\
        \cmidrule(lr){2-3}\cmidrule(lr){4-5}\cmidrule(lr){6-7}
        Train data & Dice (GT$>0$) & Detect F1 & Dice (GT$>0$) & Detect F1 & Dice (GT$>0$) & Detect F1 \\
        \midrule
        m2caiSeg (train) & 0.4477 $\pm$ 0.4477 & 0.6667 & 0.0029 $\pm$ 0.0027 & 0.0294 & 0.2485 $\pm$ 0.3785 & 0.2949 \\
        GynSurg (Video 01--08) & 0.3274 $\pm$ 0.3274 & 0.6667 & 0.8800 $\pm$ 0.0920 & 0.8529 & 0.3258 $\pm$ 0.3789 & 0.6053 \\
        Cholec80-port (train) & 0.4876 $\pm$ 0.4876 & 0.6667 & 0.6110 $\pm$ 0.0637 & 0.5588 & 0.8616 $\pm$ 0.1216 & 0.8556 \\
        Combined (cleaned) & 0.7218 $\pm$ 0.2610 & 1.0000 & 0.8185 $\pm$ 0.0840 & 0.8235 & 0.8127 $\pm$ 0.3057 & 0.8698 \\
        \bottomrule
    \end{tabular}
    \end{adjustbox}
\end{table}
\Cref{tab:results} summarizes performance across datasets. Training on Cholec80-port yields strong in-domain results on the Cholec80-port test split (Dice 0.862; Detect F1 0.856). Notably, the Cholec80-port-trained model also outperforms the m2caiSeg-trained model even when evaluated on the m2caiSeg test split, suggesting that geometrically consistent sleeve labels improve robustness beyond dataset-specific fitting.

Cross-dataset generalization to GynSurg remains challenging, likely due to domain shift in port materials, lighting, and surgical workflow. Even with a combined cleaned dataset, residual gaps indicate that visual diversity (e.g., transparent ports, boundary visibility, peripheral appearances) remains a key limiting factor.

Our cleansing of m2caiSeg indicates that many original labels are noisy under a sleeve-consistent definition; after re-annotation, only a small number of frames meet quality criteria, yielding high variance and limited generalization. Ablations without cleansing (not shown) demonstrate substantially worse transfer, supporting the conclusion that SOP-level geometric consistency is a dominant factor for cross-dataset robustness.

Failure cases typically involve (i) faint ports near the image boundary, (ii) transparent/low-contrast sleeves with visible background through the structure, and (iii) strong specular highlights obscuring sleeve boundaries.

\section{Conclusion}
We presented Cholec80-port, a high-fidelity trocar port segmentation dataset derived from Cholec80, together with a geometrically consistent annotation SOP that defines port-sleeve masks excluding the central lumen. Experiments show that models trained on Cholec80-port achieve strong robustness and improved cross-dataset generalization compared with models trained on existing public data, and that rigorous cleansing can substantially improve transfer performance.

Domain shift across datasets remains a challenge, suggesting that additional diversity in port appearances and surgical environments is necessary for universal applicability. Future work will expand port segmentation coverage and integrate port masking into geometry-based pipelines such as vSLAM, 3D reconstruction, and panoramic inference.

\pagebreak
\bibliographystyle{unsrt}
\bibliography{refs}
\end{document}